\documentclass[11pt]{article}
\usepackage[final]{acl}
\usepackage{times}
\usepackage{latexsym}
\usepackage[T1]{fontenc}
\usepackage[utf8]{inputenc}
\usepackage{microtype}
\usepackage{inconsolata}
\usepackage{graphicx}
\usepackage{booktabs}
\usepackage[table]{xcolor}
\usepackage{colortbl}
\usepackage{arydshln}
\usepackage{amsmath}
\usepackage{amssymb}
\usepackage{multirow}
\usepackage{algorithm}
\usepackage{algpseudocode}
\usepackage{array}
\usepackage[most]{tcolorbox}
\usepackage{xcolor}
\usepackage{listings}

\usepackage[most]{tcolorbox}
\usepackage{xcolor}
\usepackage{listings}

\definecolor{boxborder}{RGB}{0,45,55}
\definecolor{boxbg}{RGB}{248,248,248}

\lstdefinestyle{promptstyle}{
    basicstyle=\ttfamily\footnotesize,
    breaklines=true,
    columns=fullflexible,
    keepspaces=true,
    showstringspaces=false
}

\newtcolorbox{databox}[1]{
    width=0.98\linewidth,
    colback=boxbg,
    colframe=boxborder,
    boxrule=0.9pt,
    arc=1.5mm,
    left=7pt,
    right=7pt,
    top=5pt,
    bottom=5pt,
    title={#1},
    coltitle=white,
    colbacktitle=black,
    fonttitle=\bfseries\small,
    enhanced,
    boxed title style={
        colback=black,
        colframe=black,
        sharp corners,
        boxrule=0pt,
        left=6pt,
        right=6pt,
        top=2pt,
        bottom=2pt
    }
}

\title{ACTD: Anchor-Based Cross-Tokenizer Distillation with Residual
Regularization}

\author{
\textbf{Huiyi Zhang\textsuperscript{1},
Zijian Li\textsuperscript{1},
Xiaocheng Feng\textsuperscript{1,2}\thanks{Corresponding authors.},
Weitao Ma\textsuperscript{1},} \\
\textbf{Xiaoliang Yang\textsuperscript{1},
Yichong Huang\textsuperscript{1},
Bing Qin\textsuperscript{1,2}\footnotemark[1]} \\
\textsuperscript{1}Harbin Institute of Technology \\
\textsuperscript{2}Peng Cheng Laboratory \\
\texttt{\{hyzhang,lizijian,xcfeng,wtma,xlyang,ychuang,qinb\}@ir.hit.edu.cn}
}

\begin{document}
\maketitle

\begin{abstract}
Knowledge distillation effectively transfers reasoning capabilities from large language models to lightweight student models. To enable knowledge transfer across disparate model families, researchers increasingly explore cross-tokenizer distillation. However, cross-tokenizer distillation remains challenging due to vocabulary and sequence misalignment, while approximate vocabulary alignment can introduce additional noise into distillation. To address these challenges, we propose \textbf{A}nchor-Based \textbf{C}ross-\textbf{T}okenizer \textbf{D}istillation with Residual Regularization (\textbf{ACTD})\footnote{\url{https://github.com/hyhy892/ACTD}}. ACTD bridges structural heterogeneity through vocabulary and sequence alignment, while mitigating alignment noise via a novel anchor loss with residual regularization. We further extend this framework to a multi-teacher setting. Evaluated across five reasoning benchmarks with three distinct teacher models, ACTD achieves state-of-the-art performance. Moreover, its multi-teacher extension outperforms the strongest single-teacher and multi-teacher baselines, further demonstrating the robustness of our method.

\end{abstract}

\section{Introduction}
\begin{figure}[t]
    \centering
    \includegraphics[width=\linewidth]{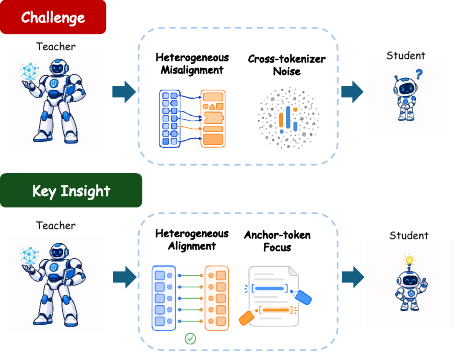}
    \caption{
    Challenges in cross-tokenizer distillation and key insight of ACTD.
    Cross-tokenizer distillation suffers from 
    (1) heterogeneous vocabulary and sequence misalignment, and 
    (2) noisy full-vocabulary supervision caused by lossy cross-tokenizer mapping. ACTD addresses these challenges by aligning heterogeneous tokenizers and focusing distillation on anchor tokens.
    }
    \label{fig:example}
\end{figure}
Large language models (LLMs) have demonstrated remarkable capabilities on reasoning tasks \citep{brown2020language,  touvron2023llama2, yang2025qwen3}. However, the real-world deployment of LLMs requires substantial computational and time costs, making knowledge distillation an essential technique for transferring the capabilities of powerful teachers to lightweight students. Compared with supervised fine-tuning (SFT) based on hard labels \citep{raffel2020exploring, wei2022finetuned}, knowledge distillation (KD) transfers soft-label supervision from teacher models \citep{hinton2015distilling}. Beyond homogeneous distillation, recent studies on cross-tokenizer distillation \citep{boizard2024uld,zhang2024dualspace,minixhofer2025universal} enable knowledge transfer between models with different tokenizers, providing greater flexibility in selecting teachers across a broader range of model families.

Despite these advantages, cross-tokenizer distillation still faces several challenges, as illustrated in Figure~\ref{fig:example}. Vocabulary mismatch \citep{boizard2024uld,zhang2024dualspace} and sequence misalignment \citep{patiño2025_unlocking_on_policy_distillation_for_any_model_family} make it difficult to align the representation spaces of teacher and student models. Beyond these alignment challenges, our analysis further shows that imperfect vocabulary alignment introduces additional noise into distillation, as discussed in Section~\ref{sec:noisy}. Furthermore, we observe that directly extending cross-tokenizer distillation methods to the multi-teacher settings leads to suboptimal performance, as alignment-induced noise can accumulate across multiple teachers. 

Motivated by these observations, we propose \textbf{Anchor-based Cross-Tokenizer Distillation (ACTD)}, a cross-tokenizer distillation method that guides the student to learn from the teacher's high-probability token distribution.
Specifically, ACTD adopts a first-subtoken vocabulary alignment strategy to map teacher tokens into the student vocabulary, and performs sequence alignment by matching nearest token positions based on character-level token boundaries. To alleviate the noise issue in cross-tokenizer distillation, ACTD optimizes the distance between teacher's high-probability tokens and student's anchor tokens through an anchor loss, and further introduces a residual regularization to suppress noise. Furthermore, we extend ACTD to the multi-teacher setting and propose \textbf{Multi-teacher Anchor-based Cross-Tokenizer Distillation (Multi-ACTD)}, demonstrating the robustness of our method.

Experimental results on five reasoning benchmarks show that ACTD improves \textsc{Avg}@8 by up to 4.5 absolute points over cross-tokenizer distillation baselines. Multi-ACTD further improves the \textsc{Avg}@8 score by 4.9 points over baselines, whereas multi-teacher extensions of other cross-tokenizer distillation methods show performance drops. This demonstrates that our method can better exploit the complementary benefits among heterogeneous teachers. Further analysis shows that both the anchor loss and residual regularization improve performance, while the residual regularization further reduces student entropy.
In summary, our contributions are as follows:
\begin{itemize}
    \item We introduce first-subtoken vocabulary alignment and nearest monotonic boundary alignment to bridge vocabulary-level and sequence-level mismatches between heterogeneous teacher and student models.
    
    \item We propose an anchor-based objective with residual regularization to alleviate noisy supervision from full-vocabulary cross-tokenizer distillation by focusing supervision on teacher high-probability tokens.

    \item We extend ACTD to the multi-teacher setting and show that Multi-ACTD consistently outperforms multi-teacher variants of existing cross-tokenizer distillation methods, demonstrating its ability to exploit complementary supervision from heterogeneous teachers and its robustness under multi-teacher distillation.
\end{itemize}
\section{Preliminaries}
Given an input sequence \(x\), a teacher model \(T\) and a student model \(S\) produce next-token distributions at each decoding position \(t\), denoted as \(p_T(\cdot \mid x_{<t})\) and \(p_S(\cdot \mid x_{<t})\), respectively. Knowledge distillation trains the student by minimizing a distributional distance between the teacher and student:
\begin{equation}
\mathcal{L}_{\mathrm{KD}}
=
\frac{1}{L}
\sum_{t=1}^{L}
D\big(
p_T(\cdot \mid x_{<t}),
p_S(\cdot \mid x_{<t})
\big),
\end{equation}
where \(D(\cdot,\cdot)\) specifies how the teacher--student discrepancy is measured, and \(L\) denotes the length of the sequence. The objective is to minimize \(\mathcal{L}_{\mathrm{KD}}\) with respect to the student-model parameters.

\paragraph{Homogeneous distillation.}
When the teacher and student share the same vocabulary \(\mathcal{V}\), their output distributions are defined over a common token space, allowing direct distribution matching. 
A common objective is the forward KL divergence \citep{hinton2015distilling}:
\begin{equation}
\mathcal{L}_{\mathrm{KL}}
=
\frac{1}{L}
\sum_{t=1}^{L}
\sum_{v \in \mathcal{V}}
p_T(v \mid x_{<t})
\log
\frac{p_T(v \mid x_{<t})}{p_S(v \mid x_{<t})} .
\end{equation}
Other objectives, including reverse KL divergence and Jensen--Shannon divergence, can also be used to control different distribution-matching behaviors \citep{gu2024minillm,agarwal2024generative}.

\paragraph{Cross-tokenizer distillation.}
When the teacher and student use different vocabularies 
\(\mathcal{V}_T\) and \(\mathcal{V}_S\), their output distributions are defined over incompatible token spaces, making token-level divergence inapplicable. 
Cross-tokenizer distillation generally introduces a cross-vocabulary transformation 
\(\Gamma: \Delta(\mathcal{V}_T) \rightarrow \Delta(\mathcal{V}_S)\), 
which projects the teacher distribution into the student vocabulary space through token mapping or distribution transport \citep{chen2025enhancing}. 
Given aligned prediction positions, the distillation objective can be written as
\begin{equation}
\mathcal{L}
=
\frac{1}{L}
\sum_{t=1}^{L}
D\big(
\Gamma(p_T(\cdot \mid x_{<t})),
p_S(\cdot \mid x_{<t})
\big),
\end{equation}
where \(D(\cdot,\cdot)\) denotes a discrepancy measure, such as KL divergence or an optimal-transport distance \citep{boizard2024uld,cui2025multilevelot}.
Since different tokenizers may also produce different sequence segmentations \citep{patiño2025_unlocking_on_policy_distillation_for_any_model_family}, cross-tokenizer distillation typically requires both sequence alignment and vocabulary alignment. 

\begin{figure*}[t]
    \centering
    \includegraphics[width=\linewidth]{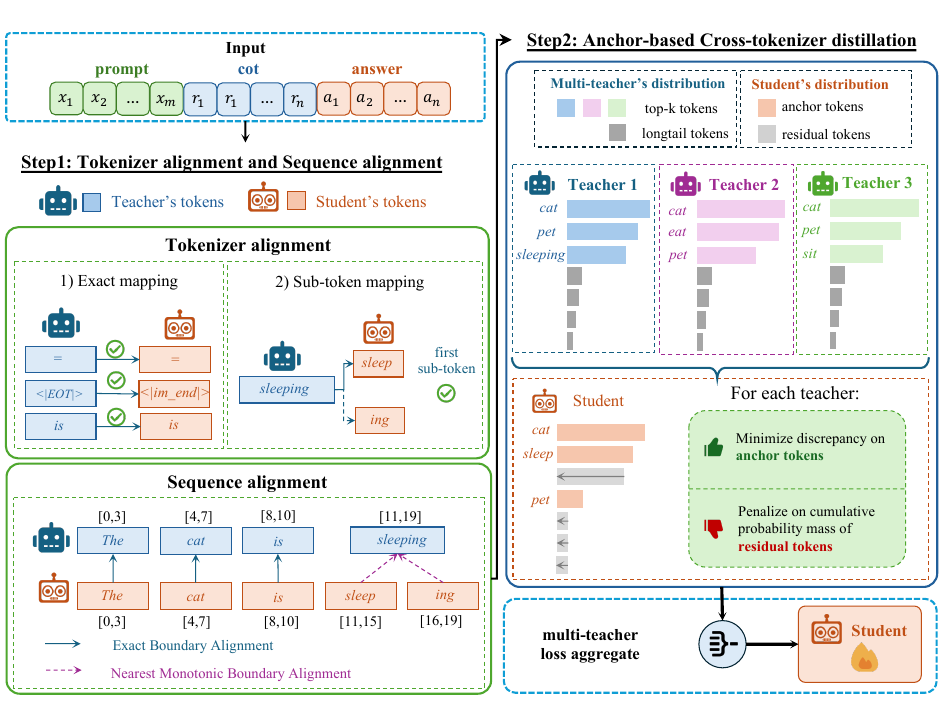}
    \caption{
    Overview of \textbf{ACTD}: 
    (1) vocabulary and sequence alignment for cross-tokenizer supervision alignment (Section~\ref{sec:alignment}); 
    (2) anchor-based loss computation with anchor loss and residual regularization for single-teacher (Sections~\ref{sec:single-teacher}) and multi-teacher distillation (Sections~\ref{sec:multi_teacher}).
    }
    \label{fig:method}
\end{figure*}
\section{Method}

ACTD consists of two main components:
(1) \textbf{Vocabulary and sequence alignment.} 
ACTD performs vocabulary alignment through first-subtoken vocabulary alignment, and conducts sequence alignment through nearest monotonic boundary alignment. (2) \textbf{Anchor-based loss with residual regularization.} 
ACTD guides the student to focus on the teacher's high-probability regions through an anchor-based loss, and introduces residual regularization to suppress noisy supervision from unsupported vocabulary regions.
For data construction, previous studies show that directly imitating complete rationales can lead to suboptimal learning \citep{liu2026long, liu2026prefixteachsuffixfade}; therefore, we construct training data with prefix-truncated rationales, with details provided in Appendix~\ref{sec:data-construction}.
\subsection{Vocabulary and Sequence Alignment}
\label{sec:alignment}
The primary challenge in cross-tokenizer distillation is to align heterogeneous supervision across both the vocabulary and sequence dimensions \citep{patiño2025_unlocking_on_policy_distillation_for_any_model_family}. Specifically, vocabulary alignment maps teacher-side token distributions into the student vocabulary space, while sequence alignment establishes correspondences between teacher and student tokens at different decoding positions.
\paragraph{Vocabulary alignment.}
\label{sec:voc-alignment}
We define a vocabulary mapping function
\begin{equation}
\Gamma: \mathcal{V}_T \rightarrow \mathcal{V}_S,
\end{equation}
which maps each teacher-side token to a student-side token. Specifically, tokens shared by the teacher and student vocabularies are directly mapped to their corresponding token ids. For unshared tokens, we first decode the teacher token into text and then re-tokenize it with the student tokenizer, using the first student subtoken as the matched token.

\paragraph{Sequence alignment.}
Cross-tokenizer distillation suffers from sequence-level misalignment \citep{patiño2025_unlocking_on_policy_distillation_for_any_model_family} because teacher and student tokenizers may segment the same text into different token sequences. 
We address this issue by aligning decoding positions through character-level token boundaries. 
For each input, we first tokenize the text with the student tokenizer and obtain student token offsets. The text is truncated to the region covered by valid student tokens and then re-tokenized with the teacher tokenizer. We convert both tokenizations into boundary positions and align each student position to the nearest teacher boundary under a monotonic constraint. The resulting position map allows teacher logits to supervise the corresponding student decoding steps.

\paragraph{Cross-tokenizer noise analysis.}
\label{sec:noisy}

\begin{figure}[t]
    \centering
    \includegraphics[width=0.95\linewidth]{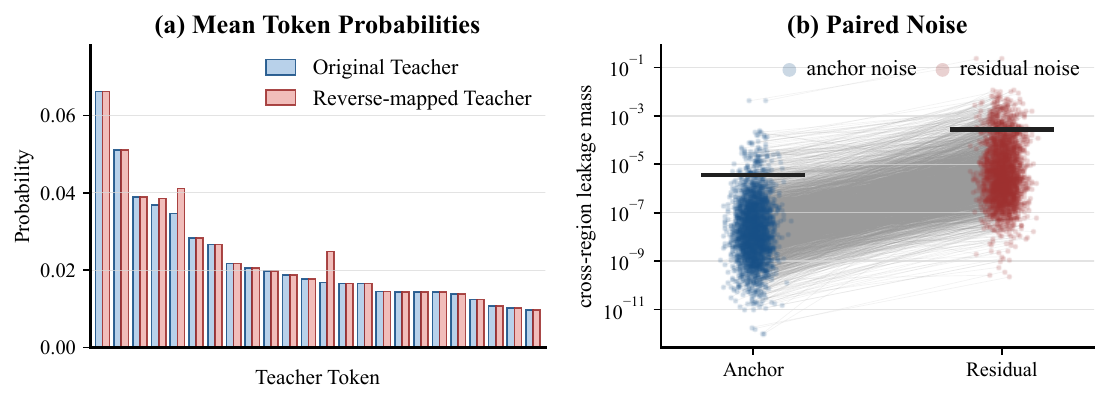}
    \caption{
    Vocabulary mapping analysis under tokenizer mismatch.
    (a) We compare the original teacher distribution \(p_T\) with the recovered
    teacher distribution after teacher-to-student mapping \(\Gamma\) and reverse
    mapping \(\Psi\), showing that vocabulary alignment redistributes probability
    mass. (b) We compare the accumulated mapped probability mass from teacher
    long-tail tokens on student-side anchor and residual regions. Anchor tokens
    are obtained by mapping teacher top-\(k\) tokens through \(\Gamma\), while
    residual tokens denote the remaining student vocabulary.
    }
    \label{fig:cross_tokenizer_noise}
\end{figure}

We first examine whether vocabulary alignment changes the teacher distribution.
Let \(\Gamma: \mathcal{V}_T \rightarrow \mathcal{V}_S\) denote the
teacher-to-student vocabulary mapping, and let
\(\Psi: \mathcal{V}_S \rightarrow \mathcal{V}_T\) denote the student-to-teacher vocabulary mapping.
Given a teacher distribution, we map its probability mass to the student
vocabulary and then map it back to the teacher vocabulary. If the alignment
were lossless, the recovered teacher distribution would match the original
teacher distribution. However, Figure~\ref{fig:cross_tokenizer_noise}(a) shows
discrepancies between the original and reverse-mapped teacher
probabilities, indicating that tokenizer alignment introduces mapping noise.

We further analyze where the mapped probability mass is concentrated in the
student vocabulary. Given the teacher distribution, we map the teacher
top-\(k\) tokens to the student
vocabulary through \(\Gamma\). These mapped tokens form the student-side
\textit{anchor tokens}, while remaining student vocabulary tokens are
defined as \textit{residual tokens}.  The detailed definition
of the noise measure for this analysis is provided in Appendix~\ref{sec:cross_tokenizer_noise}.

Figure~\ref{fig:cross_tokenizer_noise}(b)
shows that the probability mass contributed by teacher long-tail tokens
accumulates much more heavily on residual tokens than on anchor tokens.
Although each long-tail teacher token has low probability individually, their
large number leads to substantial cumulative mass after vocabulary mapping.
This suggests that the residual region receives more low-confidence mapped
teacher signal. Inspired by these observations, we propose the method described in
Section~\ref{sec:distillation_objective}.

\subsection{Residual-Regularized Anchor Distillation}
\label{sec:distillation_objective}
\subsubsection{Single-teacher Distillation}
\label{sec:single-teacher}
We first filter the teacher's full-vocabulary distribution to retain only its top-\(k\) tokens and renormalize the retained probabilities to obtain \(\mathbf{p}^{T}_{t}\). The token-level ACTD loss consists of an anchor loss and a residual loss:
\begin{equation}
\mathcal{L}_{\mathrm{ACTD}}(t)
=
\mathcal{L}_{\mathrm{anchor}}(t)
+
\lambda_{\mathrm{res}}
\mathcal{L}_{\mathrm{res}}(t),
\end{equation}
where \(t\) denotes a target position, and \(\lambda_{\mathrm{res}}\) controls the strength of residual regularization. 

The sequence-level ACTD loss is obtained by averaging the token-level loss over all target positions:
\begin{equation}
\mathcal{L}_{\mathrm{ACTD}}
=
\frac{1}{L}
\sum_{t=1}^{L}
\left(
\mathcal{L}_{\mathrm{anchor}}(t)
+
\lambda_{\mathrm{res}}
\mathcal{L}_{\mathrm{res}}(t)
\right).
\end{equation}

\paragraph{Anchor Loss.}
At target position \(t\), let \(\mathbf{p}^{T}_{t}=(p^{T}_{t,1},\ldots,p^{T}_{t,k})\) denote the teacher top-\(k\) probability distribution. The corresponding teacher-side top-\(k\) tokens are mapped to student tokens \(\mathbf{a}_{t}=(a_{t,1},\ldots,a_{t,k})\), where \(a_{t,j}\in\mathcal{V}_S\) denotes the student token mapped from the \(j\)-th teacher-side token. The anchor subset is defined as \(\mathcal{A}_{t}=\{a_{t,j}\}_{j=1}^{k}\). Let \(p^{S}_{t}(v)\) denote the student probability of token \(v\in\mathcal{V}_S\), and the student probability on the \(j\)-th anchor token is \(\hat{p}^{S}_{t,j}=p^{S}_{t}(a_{t,j})\).
The anchor loss matches the teacher top-\(k\) probability distribution with the student probabilities on the anchor subset:
\begin{equation}
\mathcal{L}_{\mathrm{anchor}}(t)
=
\sum_{j=1}^{k}
\left|p^{T}_{t,j}-\hat{p}^{S}_{t,j}\right|.
\end{equation}
This term transfers fine-grained teacher probability signals to the mapped student tokens.

\paragraph{Residual
Regularization.}
The residual subset contains all student tokens outside the anchor subset, i.e., \(\mathcal{R}_{t}=\mathcal{V}_S \setminus \mathcal{A}_{t}\). Since the teacher's long-tail probabilities are more susceptible to
cross-tokenizer alignment noise, we discard their token-level values
and set the target probabilities over the residual subset to zero.
We then constrain the student residual probability vector
\(\mathbf{p}^{S}_{t,\mathcal{R}_{t}}\) toward this zero vector, thereby
explicitly suppressing the probability mass assigned to residual tokens. We define the residual loss as the accumulated student probability mass on the residual subset:
\begin{equation}
\mathcal{L}_{\mathrm{res}}(t)
=
\left\|
\mathbf{p}^{S}_{t,\mathcal{R}_{t}}
-
\mathbf{0}
\right\|_{1}
=
\sum_{v\in \mathcal{R}_{t}} p^{S}_{t}(v).
\end{equation}

We adopt the L1 norm as the distance measure rather than the KL divergence commonly used in knowledge distillation, because residual regularization is designed to control the low-probability region rather than match a complete probability distribution. Prior work has shown that KL-based matching of low-probability regions may induce \textit{mode-averaging behavior}, resulting in an overly smoothed student distribution \citep{gu2024minillm}. In contrast, our residual regularization directly penalizes the aggregate probability mass assigned to residual tokens, thereby mitigating mode averaging and preserving a sharper student distribution.

\subsubsection{Multi-teacher Distillation}
\label{sec:multi_teacher}
We further extend ACTD to the multi-teacher setting, denoted as \textbf{Multi-ACTD}. Suppose there are \(M\) teacher models \(\{T_m\}_{m=1}^{M}\). For each teacher \(T_m\), we independently construct a cache \(\mathcal{C}^{(m)}\) following the procedure described above. During student training, Multi-ACTD retrieves the cached top-\(k\) logits from each teacher according to the same sample and target position, and computes a teacher-specific ACTD loss.

For a target position \(t\), the ACTD loss induced by teacher \(T_m\) is denoted as
\begin{equation}
\mathcal{L}^{(m)}_{\mathrm{ACTD}}(t)
=
\mathcal{L}^{(m)}_{\mathrm{anchor}}(t)
+
\lambda_{\mathrm{res}}
\mathcal{L}^{(m)}_{\mathrm{res}}(t).
\end{equation}
We aggregate the supervision from multiple teachers by averaging their teacher-specific losses:
\begin{equation}
\mathcal{L}_{\mathrm{Multi\text{-}ACTD}}(t)
=
\frac{1}{M}
\sum_{m=1}^{M}
\mathcal{L}^{(m)}_{\mathrm{ACTD}}(t).
\end{equation}
The final training objective is
\begin{equation}
\mathcal{L}_{\mathrm{Multi\text{-}ACTD}}
=
\frac{1}{|\Omega|}
\sum_{i=1}^{N}
\sum_{t\in \Omega_i}
\mathcal{L}_{\mathrm{Multi\text{-}ACTD}}(t),
\end{equation}
where \(\Omega_i\) denotes the set of supervised target positions for example \(i\), and \(|\Omega|\) is the total number of supervised target tokens.


\begin{table*}[t]
\centering
\small
\setlength{\tabcolsep}{3pt}
\resizebox{\textwidth}{!}{
\begin{tabular}{l c cc cc cc cc cc cc}
\toprule
\textbf{Model} & \textbf{Setting} 
& \multicolumn{2}{c}{AIME24} 
& \multicolumn{2}{c}{AIME25}
& \multicolumn{2}{c}{AMC23}
& \multicolumn{2}{c}{Minerva}
& \multicolumn{2}{c}{MATH500}
& \multicolumn{2}{c}{Avg} \\
\cmidrule(lr){3-4} \cmidrule(lr){5-6} \cmidrule(lr){7-8}
\cmidrule(lr){9-10} \cmidrule(lr){11-12} \cmidrule(lr){13-14}
 &  & \(\textsc{AVG}^{8}\) & \(\textsc{PASS}^{8}\)
     & \(\textsc{AVG}^{8}\) & \(\textsc{PASS}^{8}\)
     & \(\textsc{AVG}^{8}\) & \(\textsc{PASS}^{8}\)
     & \(\textsc{AVG}^{8}\) & \(\textsc{PASS}^{8}\)
     & \(\textsc{AVG}^{8}\) & \(\textsc{PASS}^{8}\)
     & \(\textsc{AVG}^{8}\) & \(\textsc{PASS}^{8}\) \\

\midrule
\multicolumn{14}{c}{
  \cellcolor{gray!15}
  \textit{Teacher Models (Reasoning under 8192-token limit)}
} \\
\midrule

DS-Llama-8B & T1 
& 26.7 & 63.3 & 21.7 & 40.0 & 73.1 & 90.0 & 32.4 & 51.1 & 81.3 & 93.4 & 47.0 & 67.6 \\

DS-Qwen-7B & T2 
& 40.8 & 73.3 & 32.5 & 56.7 & 80.6 & 92.5 & 40.7 & 53.3 & 87.9 & 95.4 & 56.5 & 74.2 \\

DS-Qwen3-8B & T3
& 6.7 & 20.0 & 33.3 & 50.0 & 30.0 & 42.5 & 33.0 & 41.9 & 66.1 & 77.0 & 33.8 & 46.3 \\ 

\midrule
\multicolumn{14}{c}{
  \cellcolor{gray!15}
  \textit{Baselines}
} \\
\midrule

Qwen3-1.7B-Base & S 
& 2.1 & 16.7 & 1.3 & 6.7 & 27.5 & 67.5 & 17.2 & 41.5 & 51.8 & 83.2 & 20.0 & 43.1 \\

SFT & - 
& 5.0 & 16.7 & 3.3 & 16.7 & 27.2 & 62.5 & 20.0 & 43.8 & 57.1 & 81.8 & 22.5 & 44.3 \\

KD & Homo.
& 8.3 & 23.3 & 4.6 & 16.7 & 34.7 & 72.5 & \textbf{25.1} & 45.2 & 60.8 & 85.8 & 26.7 & 48.7\\

\midrule
\multicolumn{14}{c}{
  \cellcolor{gray!15}
  \textit{Single-teacher}
} \\
\midrule

DSKD & T1$\to$S
& 5.0 & 13.3 & 2.9 & 13.3 & 29.4 & 62.5 & 22.0 & 43.0 & 57.4 & 81.8 & 23.3 & 42.8 \\

ALM & T1$\to$S
& 5.0 & 20.0 & 1.7 & 6.7 & 29.7 & 72.5 & 18.4 & 41.9 & 53.5 & 82.4 & 21.6 & 44.7  \\

ULD & T1$\to$S
& 4.6 & 13.3 & 3.8 & 16.7 & 30.3 & 62.5 & 18.4 & 38.2 & 57.4 & 83.4 & 22.9 & 42.8 \\

\cellcolor{blue!10}ACTD(Ours) & T1$\to$S
& \underline{9.2} & \underline{26.7} & \underline{5.0} & \underline{20.0} & \underline{40.0} & \underline{77.5} & \underline{22.8} & \underline{44.1} & \underline{62.0} & \underline{85.4} & \underline{27.8} & \underline{50.7} \\

\\[-0.65em]
\cdashline{1-14}
\\[-0.55em]

DSKD & T2$\to$S
& 3.8 & 20.0 & 2.5 & 13.3 & 31.9 & 65.0 & 20.2 & \underline{43.8} & 57.1 & 84.2 & 23.1 & 45.3  \\

ALM & T2$\to$S
& 4.2 & 13.3 & 1.3 & 10.0 & 25.9 & 57.5 & 18.4 & 40.4 & 53.4 & 82.0 & 20.6 & 40.7 \\

ULD & T2$\to$S 
& \underline{7.1} & \underline{23.3} & 3.3 & 13.3 & 33.4 & 62.5 & 19.7 & 41.5 & 57.4 & 83.0 & 24.2 & 44.8 \\

\cellcolor{blue!10}ACTD(Ours) & T2$\to$S 
& \underline{7.1} & 20.0 & \underline{5.4} & \underline{20.0} & \underline{35.6} & \underline{75.0} & \underline{24.8} & 43.4 & \underline{61.4} & \underline{85.8} & \underline{26.9} & \underline{48.8} \\

\\[-0.65em]
\cdashline{1-14}
\\[-0.55em]

DSKD & T3$\to$S
& 5.4 & \underline{23.3} & 2.1 & 10.0 & 27.8 & 65.0 & 19.7 & 43.4 & 49.3 & 81.6 & 20.9 & 44.7 \\

ALM & T3$\to$S
& 3.8 & 13.3 & 1.3 & 6.7 & 28.4 & 57.5 & 19.3 & \underline{43.8} & 55.6 & 83.0 & 21.6 & 40.9  \\

ULD & T3$\to$S 
& 5.0 & 16.7 & 3.8 & 16.7 & 33.8 & 62.5 & \underline{21.3} & 43.4 & 58.9 & 84.0 & 24.5 & 44.6 \\

\cellcolor{blue!10}ACTD(Ours) & T3$\to$S
& \underline{7.9} & 20.0 & \underline{5.4} & \underline{16.7} & \underline{38.8} & \underline{67.5} & 21.1 & 43.0 & \underline{62.1} & \underline{84.8} & \underline{27.1} & \underline{46.4} \\

\midrule
\multicolumn{14}{c}{
  \cellcolor{gray!15}
  \textit{Multi-Teacher}
} \\
\midrule

SeqKD & T1+T2+T3$\to$S 
& 4.6 & 16.7 & 3.3 & 13.3 & 27.8 & 70.0 & 17.3 & 39.3 & 51.6 & 82.2 & 20.9 & 44.3 \\

Multi-ULD & T1+T2+T3$\to$S 
& 4.6 & 16.7 & 5.0 & 13.3 & 30.6 & 57.5 & 21.6 & 40.8 & 58.8 & 82.0 & 24.1 & 42.1 \\

\cellcolor{blue!10}Multi-ACTD(Ours) & T1+T2+T3$\to$S 
& \textbf{10.0} & \textbf{26.7} & \textbf{6.3} & \textbf{20.0} & \textbf{40.9} & \textbf{77.5} & 24.0 & \textbf{45.6} & \textbf{63.6} & \textbf{85.8} & \textbf{29.0} & \textbf{51.1} \\

\bottomrule
\end{tabular}
}
\caption{
Comparison of results for LLM reasoning ability distillation. 
We use \(T\to S\), ``Homo.'', and ``DS'' to denote teacher-to-student distillation, the homogeneous teacher setting, and DeepSeek-series distilled models, respectively. 
ACTD is compared with other cross-tokenizer distillation methods under the same teacher setting, and the best results in each such group are marked in \underline{underline}. 
Multi-ACTD is compared with all student baselines, and the best results are marked with \textbf{bold}. 
\textsc{Avg}$^{8}$ and \textsc{Pass}$^{8}$ denote \textsc{Avg}@8 and \textsc{Pass}@8, respectively.
}
\label{tab:main_results}
\end{table*}

\section{Experiments}
\subsection{Experimental Setup}

\paragraph{Training Settings.}
We use DeepSeek-R1-Distill-Qwen-7B, DeepSeek-R1-Distill-Llama-8B, and DeepSeek-R1-0528-Qwen3-8B as teacher models, and Qwen3-1.7B-Base as the student model \citep{deepseekai2025deepseekr1, deepseek2025r10528qwen3, yang2025qwen3}. 
The tokenizer heterogeneity between the teacher and student models is analyzed in Appendix~\ref{sec:tokenizer_heterogeneity}. 
To further evaluate the generalization of ACTD across model families, we additionally use GLM-4-9B-Chat as another teacher model \citep{glm2024chatglm}.

For training data, we randomly sample 30,000 examples from the math domain of OpenThoughts-114k \citep{guha2025openthoughts}. Following recent KD methods \citep{jung2025todi, wang2025abkd}, we adopt a standard off-policy distillation setup, using the \texttt{problem} field as input and the dataset-provided \texttt{ground\_truth\_solution} as the target sequence.  For each teacher model, we run inference on the training set and cache the top-\(k\) logits with \(k=128\) for distillation. 
\paragraph{Baselines.}
We compare our method with both single-teacher cross-tokenizer distillation baselines and multi-teacher distillation baselines.

For cross-tokenizer distillation baselines:
\begin{itemize}
    \item \textbf{SFT} \citep{raffel2020exploring, wei2022finetuned}: 
    We optimize the student on the training text data with the standard cross-entropy loss.

    \item \textbf{KD} \citep{hinton2015distilling}: 
    We use the tokenizer-compatible Qwen3-8B as the homogeneous teacher and Qwen3-1.7B-Base as the student, consistent with the other baselines.

    \item \textbf{ULD} \citep{boizard2024uld}: 
    Universal Logit Distillation uses an optimal-transport-based objective to enable logits-level distillation between models with different tokenizers.

    \item \textbf{DSKD} \citep{zhang2024dualspace}: 
    Dual-Space Knowledge Distillation unifies teacher and student output spaces and introduces cross-model attention for knowledge transfer.

    \item \textbf{ALM} \citep{minixhofer2025universal}: 
    Approximate Likelihood Matching performs cross-tokenizer distillation by approximately matching across different tokenizations.
\end{itemize}

For multi-teacher distillation baselines:
\begin{itemize}
    \item \textbf{SeqKD} \citep{kim2016sequence}: 
    Sequence-level Knowledge Distillation trains the student on teacher-generated outputs rather than token-level soft labels. In the multi-teacher setting, we collect outputs from all teachers, shuffle the combined data, and fine-tune the student on the resulting mixture.

    \item \textbf{Multi-ULD}: 
    We extend ULD to the multi-teacher setting by computing the ULD loss for each teacher independently and averaging the teacher-specific losses as the final objective.
\end{itemize}

\paragraph{Evaluation Settings.}
We evaluate our method on a broad range of mathematical reasoning tasks. AIME24, AIME25, and AMC23 are competition-style benchmarks designed to assess advanced mathematical problem-solving ability \citep{yao2025varmath}. MATH500 is a representative 500-problem subset of the MATH benchmark \citep{hendrycks2021math}, following the evaluation split used in process-supervision research \citep{lightman2023verify}. Minerva consists of college-level quantitative reasoning problems across mathematics and scientific domains \citep{lewkowycz2022minerva}. We report accuracy as the main metric. 

For inference, we use the same decoding configuration for all methods:
\(\texttt{max\_gen\_tokens}=8192\), \(\texttt{temperature}=1.0\), and \(\texttt{top\_p}=0.8\).
 Following the evaluation protocol of OPSD
\citep{zhao2026self}, we extract the answer enclosed in
\(\backslash\texttt{boxed}\{\}\) from each generated response and use the
\texttt{math\_verify}\footnote{\url{https://github.com/huggingface/Math-Verify}}
package to parse both the extracted answer and the ground-truth answer and
determine their mathematical equivalence. Accuracy is then computed as the
proportion of generated responses judged correct.
\section{Results}
\subsection{Main Results}
\paragraph{Cross-tokenizer distillation results.}
Table~\ref{tab:main_results} shows that ACTD consistently outperforms cross-tokenizer distillation baselines under the same teacher settings. Compared with the strongest single-teacher distillation baseline, Multi-ACTD yields consistent and substantial improvements, indicating that the student can benefit from complementary supervision provided by multiple heterogeneous teachers. Moreover, Multi-ACTD also outperforms the multi-teacher baseline, demonstrating the robustness of our method.

\paragraph{Student Models from Other Families.}
To further evaluate the generalization of ACTD across different student architectures and model scales, we conduct additional experiments using DeepSeek-R1-Distill-Qwen-7B as the teacher and Llama-3.2-3B-Instruct as the student. This setting introduces both a different student architecture and a larger model scale compared with our primary experiments based on Qwen3-1.7B-Base. As shown in Table~\ref{tab:llama_student_results}, ACTD achieves the best overall performance, obtaining an average \textsc{Avg}@8 score of 18.4 across five mathematical reasoning benchmarks. Compared with the strongest baseline, DSKD, which achieves 15.8, ACTD provides an absolute improvement of 2.6 points. ACTD also achieves the best performance on four benchmarks and ties for the best result on AIME25. These results demonstrate that ACTD generalizes effectively across different student architectures and model scales.

\paragraph{Teacher Models from Other Families.}
To further evaluate the robustness of ACTD beyond the DeepSeek family, we conducted a single-teacher distillation using GLM-4-9B-Chat as the teacher and Qwen3-1.7B-Base as the student. As shown in Table~\ref{tab:glm_single_teacher_results}, ACTD outperforms the cross-tokenizer distillation baselines on mathematical reasoning benchmarks. ACTD improves \textsc{Avg}@8 from 22.7 to 26.0 and \textsc{Pass}@8 from 44.9 to 47.2 compared with the strongest baseline. These results indicate that ACTD remains effective in other model families, demonstrating the robustness of our approach.

\begin{table}[t]
    \centering
    \small
    \setlength{\tabcolsep}{3pt}
    \resizebox{\columnwidth}{!}{
    \begin{tabular}{lcccccc}
        \toprule
        \textbf{Method}
        & \textbf{AIME24}
        & \textbf{AIME25}
        & \textbf{AMC23}
        & \textbf{Minerva}
        & \textbf{MATH500}
        & \textbf{Average} \\
        \midrule
        Llama-3.2-3B-Instruct
        & 0.4 & 0.4 & 15.0 & 12.5 & 31.9 & 12.0 \\
        ALM
        & 2.5 & \textbf{1.3} & 20.0 & 11.2 & 40.9 & 15.2 \\
        DSKD
        & 3.8 & \textbf{1.3} & 20.6 & 11.6 & 41.6 & 15.8 \\
        ULD
        & 4.2 & 0.4 & 13.4 & 12.4 & 29.9 & 12.1 \\
        \textbf{ACTD}
        & \textbf{5.8}
        & \textbf{1.3}
        & \textbf{23.1}
        & \textbf{16.3}
        & \textbf{45.5}
        & \textbf{18.4} \\
        \bottomrule
    \end{tabular}
    }
    \caption{
        Single-teacher distillation results under the
        DeepSeek-R1-Distill-Qwen-7B $\rightarrow$ Llama-3.2-3B-Instruct setting.
        All results are reported in \textsc{Avg}@8.
        The best results are highlighted in bold.
    }
    \label{tab:llama_student_results}
\end{table}

\subsection{Ablation Studies}
\begin{figure}[t]
    \centering
    \includegraphics[width=0.95\linewidth]{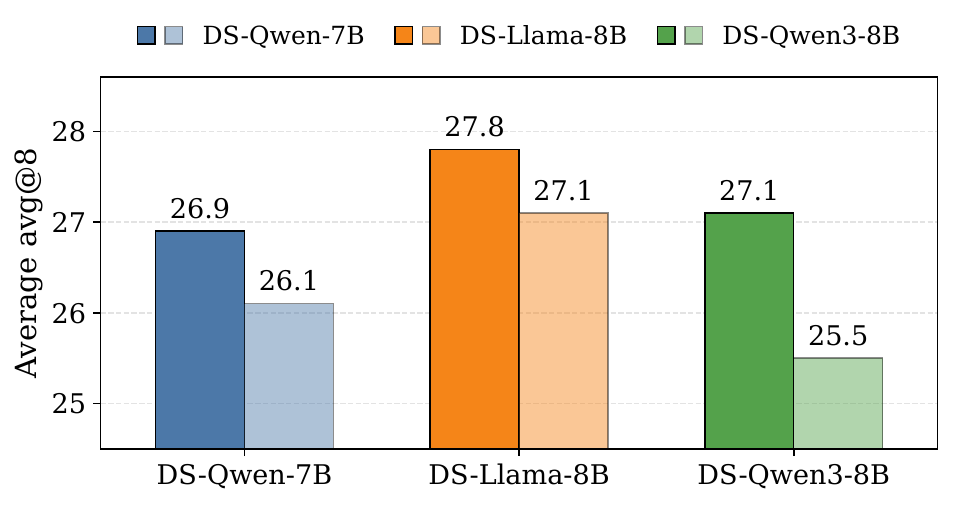}
    \caption{Effect of prefix-truncated reasoning targets. The figure reports the average performance on mathematical reasoning benchmarks under different teacher settings.
    Light-colored bars indicate the results after removing prefix truncation.
    }
    \label{fig:truncation_ablation}
\end{figure}

\begin{figure}[t]
    \centering
    \includegraphics[width=0.95\linewidth]{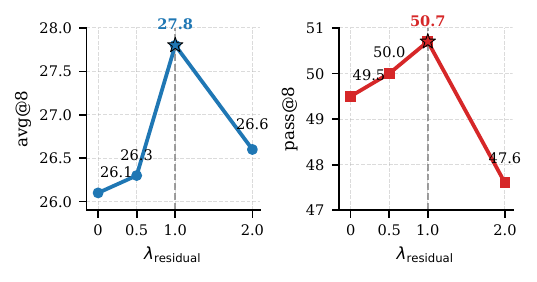}
    \caption{Effect of the Parameter \(\lambda_{\mathrm{res}}\). The figure reports the average \textsc{Avg}@8 and \textsc{Pass}@8 scores across all mathematical reasoning benchmarks.}
    \label{fig:lambda_res_ablation}
\end{figure}

\paragraph{Effect of Prefix-Truncated Reasoning Targets.}
We first conduct an ablation study to evaluate the effectiveness of prefix-truncated reasoning target construction. 
As shown in Figure~\ref{fig:truncation_ablation}, removing prefix truncation consistently leads to a performance drop of 0.7--1.6 points in \textsc{Avg}@8 across different teacher settings. 
Notably, the degradation is most pronounced when using DeepSeek-R1-Distill-Qwen3-8B, a teacher with stronger long-reasoning capabilities. 
This suggests that prefix-truncated targets facilitate more effective knowledge transfer under constrained training context lengths.
\begin{table}[t]
\centering
\small
\setlength{\tabcolsep}{3pt}
\resizebox{\linewidth}{!}{
\begin{tabular}{lcccccc}
\toprule
\multicolumn{7}{c}{\textbf{GLM-4-9B-Chat} $\rightarrow$ \textbf{Qwen3-1.7B-Base}} \\
\midrule
\multicolumn{7}{c}{\textsc{Avg}@8} \\
\midrule
\textbf{Method} & \textbf{AIME24} & \textbf{AIME25} & \textbf{AMC23} & \textbf{Minerva} & \textbf{MATH500} & \textbf{Avg.} \\
\midrule
ULD            & 2.5 & 2.1 & 29.7 & 17.1 & 53.5 & 21.0 \\
ALM            & 3.3 & 2.1 & 25.9 & 20.1 & 53.2 & 20.9 \\
DSKD           & 3.3 & 2.9 & 30.0 & 19.9 & 57.4 & 22.7 \\
\textbf{ACTD}      & \textbf{10.0} & \textbf{3.8} & \textbf{33.4} & \textbf{23.7} & \textbf{59.3} & \textbf{26.0} \\
\midrule
\multicolumn{7}{c}{\textsc{Pass}@8} \\
\midrule
\textbf{Method} & \textbf{AIME24} & \textbf{AIME25} & \textbf{AMC23} & \textbf{Minerva} & \textbf{MATH500} & \textbf{Avg.} \\
\midrule
ULD            & 13.3 & 10.0 & 65.0 & 40.8 & 82.8 & 42.4 \\
ALM            & 16.7 & \textbf{16.7} & 67.5 & 43.0 & 80.8 & 44.9 \\
DSKD           & 16.7 & 10.0 & 62.5 & \textbf{44.5} & 81.8 & 43.1 \\
\textbf{ACTD}      & \textbf{23.3} & 13.3 & \textbf{72.5} & 43.0 & \textbf{83.8} & \textbf{47.2} \\
\bottomrule
\end{tabular}
}
\caption{Single-teacher distillation results under the GLM-4-9B-Chat $\rightarrow$ Qwen3-1.7B-Base setting. ACTD achieves the best overall performance among cross-tokenizer distillation baselines.}
\label{tab:glm_single_teacher_results}
\end{table}
\paragraph{Effect of the Parameter \(\lambda_{\mathrm{res}}\).}
We further study the sensitivity of ACTD to the residual loss weight \(\lambda_{\mathrm{res}}\). We vary \(\lambda_{\mathrm{res}}\) over \(\{0, 0.5, 1.0, 2.0\}\) and evaluate on mathematical reasoning benchmarks. Figure~\ref{fig:lambda_res_ablation} shows the average \textsc{Avg}@8 and \textsc{Pass}@8 scores across all tasks.

The results reveal two trends. 
(1) \textbf{Low \(\lambda_{\mathrm{res}}\).} Setting \(\lambda_{\mathrm{res}}=0\) and \(\lambda_{\mathrm{res}}=0.5\) decreases \textsc{Avg}@8 to 26.1 and 26.3, and decreases \textsc{Pass}@8 to 49.5 and 50.0, respectively, indicating that residual regularization is important for suppressing excessive probability mass on non-anchor tokens. 
(2) \textbf{High \(\lambda_{\mathrm{res}}\).} At \(\lambda_{\mathrm{res}}=2.0\), \textsc{Avg}@8 drops to 26.6 and \textsc{Pass}@8 drops to 47.6, suggesting that an overly strong residual penalty may weaken anchor loss and reduce the transfer of informative teacher top-\(k\) signals. 
Overall, \(\lambda_{\mathrm{res}}=1.0\) achieves the best balance, highlighting the complementary roles of anchor loss and residual loss.

\paragraph{Effect of the Top-\(k\) Size.}
\begin{table}[t]
\centering
\small
\setlength{\tabcolsep}{3pt}
\resizebox{\linewidth}{!}{
\begin{tabular}{lcccccc}
\toprule
\textbf{\(k\)} 
& \textbf{AIME24} 
& \textbf{AIME25} 
& \textbf{AMC23} 
& \textbf{Minerva} 
& \textbf{MATH500} 
& \textbf{Avg} \\
\midrule
64  & 8.3 & 5.0 & 34.7 & 22.4 & 62.8 & 26.6 \\
96  & 9.2 & 5.4 & 35.9 & 23.0 & 60.1 & 26.7 \\
128 & 9.2 & 5.0 & 40.0 & 22.8 & 62.0 & \textbf{27.8} \\
256 & 8.3 & 5.8 & 36.9 & 22.0 & 60.0 & 26.6 \\
full vocab. & 9.2 & 5.8 & 35.9 & 20.8 & 58.0 & 25.9 \\
\bottomrule
\end{tabular}
}
\caption{Effect of the number of teacher top-\(k\) logits. The table reports \textsc{Avg}@8 scores under the DeepSeek-R1-Distill-Llama-8B \(\rightarrow\) Qwen3-1.7B-Base setting.}
\label{tab:topk_ablation}
\end{table}

As shown in Table~\ref{tab:topk_ablation}, we further analyze the effect of the number of teacher logits by varying \(k\) over \(\{64,96,128,256,\text{full vocab}\}\) under the DeepSeek-R1-Distill-Llama-8B \(\rightarrow\) Qwen3-1.7B-Base setting. 
The table reports the \textsc{Avg}@8 scores across mathematical reasoning benchmarks. The student achieves the best average performance when \(k=128\), reaching 27.8 on average.

When \(k\) is further increased, the average \textsc{Avg}@8 score drops to 26.6. This indicates that simply expanding the teacher top-\(k\) supervision does not necessarily improve distillation. Instead, including more low-probability long-tail tokens introduce additional noisy supervision, which weakens the precision of knowledge transfer. These results support our motivation of focusing distillation on a compact set of teacher logits rather than broadly matching larger vocabulary-level distributions.

\begin{figure}[t]
    \centering
    \includegraphics[width=\linewidth]{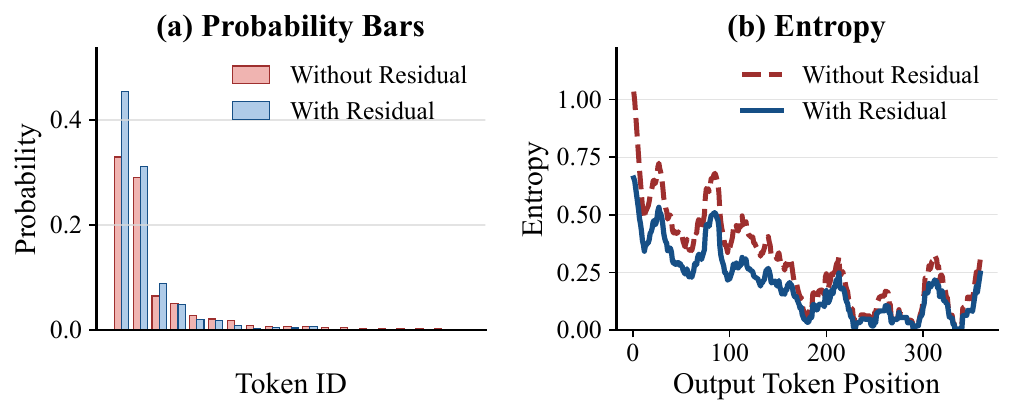}
    \caption{
    Comparison of student models distilled with and without the residual regularization. 
    The figure shows the changes in (a) probability distribution and (b) entropy across inference steps. 
    }
    \label{fig:residual-analysis}
\end{figure}

\subsection{Effect of the Residual Regularization}
\label{sec:residual-analysis}

To further analyze the effect of the residual loss, we compare two student models trained with different residual weights: one with the residual regularization (\(\lambda_{\mathrm{res}}=1\)) and the other without the residual regularization (\(\lambda_{\mathrm{res}}=0\)). 
The residual loss is designed to regularize the probability mass assigned to non-anchor regions, encouraging the student model to concentrate more on the teacher model's high-probability tokens.

As shown in Figure~\ref{fig:residual-analysis}, the model trained with the residual regularization exhibits a more concentrated probability distribution during inference. 
Specifically, Figure~\ref{fig:residual-analysis} (a) shows that the student with residual regularization assigns more probability mass to high-probability regions, indicating that the residual loss helps reduce over-smoothed predictions. 
Moreover, Figure~\ref{fig:residual-analysis} (b) shows that the residual-constrained student consistently has lower entropy across inference steps, indicating that its output distribution is less dispersed over the vocabulary. 
This suggests that the residual loss suppresses probability spreading over non-anchor regions and mitigates the over-smoothing effect caused by full-vocabulary distillation. 
These results demonstrate that the residual regularization effectively improves the concentration of the student distribution and helps preserve the sharp structure of teacher supervision.
\section{Related Work}
\subsection{Single-teacher Distillation}

Knowledge distillation (KD) transfers teacher knowledge by aligning teacher and student predictive distributions with objectives such as KL, reverse KL, or Jensen--Shannon divergence \citep{hinton2015distilling,gu2024minillm, wen2023f}. 
However, the shared-vocabulary assumption limits their applicability across model families.

Cross-tokenizer distillation relaxes the shared-vocabulary assumption by transferring distribution-level supervision across heterogeneous token spaces. 
Recent methods address tokenizer mismatch through optimal transport \citep{boizard2024uld,cui2025multilevelot}, dual-space alignment with cross-model attention \citep{zhang2024dualspace}, approximate likelihood matching \citep{minixhofer2025universal}, or span-level preference projection \citep{nguyen2026ctpd}. 
Despite their effectiveness, these methods mainly focus on single-teacher transfer and do not directly address efficient logits-level distillation from multiple heterogeneous teachers.

\subsection{Multi-Teacher Distillation}

Multi-teacher distillation improves student learning by leveraging complementary supervision from multiple teachers.
Existing methods mainly supervise student learning with teacher-generated responses or Chain-of-Thought rationales, including teacher-forcing CoT \citep{tian2025beyond}, multiple-rationale consistency \citep{chen2023mcckd}, generative distribution fusion \citep{wan2024knowledgefusion}, domain-adaptive data selection \citep{liu2024ddk}, and efficient reasoning behavior distillation \citep{xu2025twt}. 
Since different teachers may provide inconsistent distributions, rationales, or reasoning preferences, recent studies further explore knowledge purification and compatibility-aware teacher fusion to mitigate conflicting supervision \citep{jin2026purification,cui2026compact}. 
More recently, multi-teacher distillation has been applied to multi-domain learning by employing domain-specialized teachers \citep{xiaomi2026mimov2flash,deepseekai2026deepseekv4highlyefficientmilliontoken}.
However, these teachers are homogeneous with the student in terms of model architecture and tokenizer, and must be separately post-trained on domain-specific data, resulting in substantial additional training costs.
In contrast, multi-teacher distillation from heterogeneous teachers across different model families and tokenizers remains largely unexplored.
\section{Conclusion}

In this paper, we propose Anchor-based Cross-Tokenizer Distillation (ACTD) for heterogeneous teacher--student pairs. ACTD first aligns their vocabularies through first-subtoken mapping and their sequences through nearest-position matching based on character-level boundaries. It then distills aligned high-probability predictions with an anchor loss and applies residual regularization to suppress probability mass outside the anchor subset, mitigating alignment noise and mode averaging.

We further extend ACTD to Multi-ACTD, a multi-teacher framework that integrates logits-level supervision from multiple heterogeneous teachers. Experiments on reasoning benchmarks show that ACTD outperforms existing cross-tokenizer distillation methods, and Multi-ACTD further improves performance by leveraging complementary teacher supervision. These results demonstrate the robustness of our method across different teacher settings and tokenizer heterogeneity levels. Looking forward, ACTD provides a flexible way to transfer logits-level knowledge across model families, which may support broader and more efficient distillation from diverse heterogeneous teachers.

\section*{Limitations}
Although ACTD effectively enables cross-tokenizer distillation, it requires access to teacher predictive distributions during data construction, which limits its applicability to models with accessible outputs, such as open-source models. While ACTD shows consistent gains across reasoning benchmarks and teacher model families, our experiments are constrained by computational resources, especially for multi-teacher combinations. Future work could extend ACTD to broader heterogeneous teacher--student settings and improve the efficiency of leveraging distributional supervision from diverse large language models.

\section*{Acknowledgements}
Xiaocheng Feng and Bing Qin are the co-corresponding authors of this work. We thank the anonymous reviewers for their insightful comments. This work was supported by the National Natural Science Foundation of China (NSFC) (grant 62522603, 62276078, U22B2059), the Key R\&D Program of Heilongjiang via grant 2022ZX01A32, and the Fundamental Research Funds for the Central Universities ( XNJKKGYDJ2024013 ).


\bibliography{main}

\appendix
\clearpage

\section{Tokenizer Heterogeneity}
\label{sec:tokenizer_heterogeneity}
To clarify the tokenizer relations in our experiments, we summarize the vocabulary differences between each teacher model and the student model in Table~\ref{tab:tokenizer_heterogeneity}. Our single-teacher experiments evaluate ACTD under strong tokenizer heterogeneity, while our multi-teacher experiments study whether ACTD can be extended to teachers with varying degrees of tokenizer mismatch. Specifically, DeepSeek-R1-Distill-Llama-8B uses a different base vocabulary from Qwen3-1.7B-Base, while DeepSeek-R1-Distill-Qwen-7B and DeepSeek-R1-0528-Qwen3-8B share the same base vocabulary with the student but differ in tokenizer-specific special tokens. This mixed-tokenizer setting reflects practical multi-teacher distillation scenarios, where teacher quality and tokenizer compatibility jointly affect knowledge transfer.

\section{Cross-tokenizer Noise Analysis}
\label{sec:cross_tokenizer_noise}
\paragraph{Noise definition.}
For each target position, let \(\Gamma: \mathcal{V}_T \rightarrow \mathcal{V}_S\)
be the teacher-to-student vocabulary mapping. We define the student-side anchor
set by mapping the teacher top-\(k\) tokens:
\begin{equation}
\mathcal{A}_S=\{\Gamma(v)\mid v\in\operatorname{TopK}(p_T)\}.
\end{equation}
The residual set is the remaining student vocabulary:
\begin{equation}
\mathcal{R}_S=\mathcal{V}_S\setminus\mathcal{A}_S.
\end{equation}
We then measure how much mapped teacher probability mass from non-top-\(k\)
teacher tokens falls into these two regions. The accumulated mass on
\(\mathcal{A}_S\) is anchor noise, while the accumulated mass on
\(\mathcal{R}_S\) is residual noise.

\paragraph{Student-side mapping noise.}
We define vocabulary mapping noise in the student vocabulary space.
For each target position $t$, the teacher model first produces a probability
distribution over its own vocabulary,
\begin{equation}
p_T^t(v), \qquad v \in \mathcal{V}_T .
\end{equation}
This distribution is used only as the source probability mass. We then project
it onto the student vocabulary using the teacher-to-student tokenizer alignment
map \(f: \mathcal{V}_T \rightarrow \mathcal{V}_S\):
\begin{equation}
q_{\mathrm{full}}^t(s)
=
\sum_{v \in \mathcal{V}_T: f(v)=s} p_T^t(v),
\qquad s \in \mathcal{V}_S .
\end{equation}
Thus, all noise quantities below are computed over student tokens \(s\), not
teacher tokens \(v\).

Let \(\mathcal{A}_T^t=\operatorname{TopK}(p_T^t)\) denote the teacher top-\(K\)
tokens at position \(t\). We map these top-\(K\) tokens to the student vocabulary
and obtain the student-side anchor set
\begin{equation}
\mathcal{A}_S^t
=
\{ f(v) \mid v \in \mathcal{A}_T^t \}.
\end{equation}
The remaining student tokens form the student-side residual set
\begin{equation}
\mathcal{R}_S^t
=
\mathcal{V}_S \setminus \mathcal{A}_S^t .
\end{equation}

To isolate the noise introduced by non-top-\(K\) teacher tokens, we also
construct a clean top-\(K\)-only projection:
\begin{equation}
q_{\mathrm{top}K}^t(s)
=
\sum_{v \in \mathcal{A}_T^t: f(v)=s} p_T^t(v),
\qquad s \in \mathcal{V}_S .
\end{equation}
The student-side mapping discrepancy is then
\begin{equation}
\Delta^t(s)
=
\left|
q_{\mathrm{full}}^t(s)
-
q_{\mathrm{top}K}^t(s)
\right|,
\qquad s \in \mathcal{V}_S .
\end{equation}

We define anchor noise as the discrepancy accumulated on student-side anchor
tokens:
\begin{equation}
\mathcal{N}_{\mathrm{anchor}}^t
=
\sum_{s \in \mathcal{A}_S^t}
\Delta^t(s),
\end{equation}
and residual noise as the discrepancy accumulated on student-side residual
tokens:
\begin{equation}
\mathcal{N}_{\mathrm{residual}}^t
=
\sum_{s \in \mathcal{R}_S^t}
\Delta^t(s).
\end{equation}
We report these quantities across all valid target positions. This analysis therefore measures how teacher probability mass, after being projected into the student vocabulary, enters the student-side anchor and residual regions. We operationally define the resulting excess mass as projection noise because it lies outside the retained high-confidence teacher support but still contributes to the student-side distillation target, thereby introducing potentially unintended supervision.

\begin{table}[t]
\centering
\small
\setlength{\tabcolsep}{4pt}
\begin{tabular}{lcc}
\toprule
\textbf{Teacher} & \textbf{Base Vocab.} & \textbf{Special Tokens} \\
\midrule
GLM-4-9B & Different & Different \\
DS-Llama-8B & Different & Different \\
DS-Qwen-7B & Same & Different \\
DS-Qwen3-8B & Same & Different \\
\bottomrule
\end{tabular}
\caption{Tokenizer relations between teacher models and the student Qwen3-1.7B-Base. DS denotes DeepSeek-R1-Distill.}
\label{tab:tokenizer_heterogeneity}
\end{table}

\section{Data Construction}
\label{sec:data-construction}

We construct prefix-truncated training data for ACTD from a math instruction dataset, where each example contains an \texttt{id}, an \texttt{instruction}, and a reference \texttt{output}. 
Each instruction is first converted into a chat-style prompt using the student tokenizer template. 
We then append a fixed formatting instruction: ``Please reason step by step, and put your final answer within \texttt{\textbackslash boxed\{\}}'' and record the token position where the assistant completion begins. 
The training sequence is formed by concatenating the formatted prompt and the reference solution.

As illustrated in Figure~\ref{fig:prefix_truncation_example}, the key step is to truncate only the reasoning prefix while preserving the final answer. 
For each completion, we split the text into a reasoning prefix and an answer suffix. 
If the completion contains a natural \texttt{</think>} marker, the text before the marker is treated as the reasoning prefix and the remaining text as the answer suffix. 
Otherwise, we locate the final answer by searching for the last \texttt{\textbackslash boxed\{\}} expression; if no boxed answer is found, we use the last sentence as a fallback suffix. 
The reasoning prefix is capped at 2,000 student tokens. 
When the prefix exceeds this budget, we move the truncation point back to the nearest sentence boundary within a 600-character search window, so that the truncated prefix remains coherent. 
Finally, we insert a forced \texttt{</think>} marker between the truncated reasoning prefix and the preserved answer suffix.

\section{Implementation Details}
\subsection{Baseline training details}
\label{app:training-hparams}

To ensure a fair comparison between DSKD, ULD and ALM, we keep the main training
configuration identical for the three baselines. Table~\ref{tab:shared-training-hparams}
summarizes the shared hyperparameters used in our implementation. We only report the
training settings that directly affect optimization, memory usage, and distillation
strength, while omitting method-specific options and dataset/model information.

All three baselines are trained with full-parameter fine-tuning under the same FSDP2
distributed training backend. We use two GPUs with bfloat16 precision and enable
gradient checkpointing to reduce memory consumption. The global batch size is set to
36, with a micro-batch size of 3 per GPU and 6 gradient accumulation steps. For
optimization, we train for 300 steps using a learning rate of \(1\times10^{-5}\) and a
warmup ratio of 0.1. The distillation loss weight is fixed to 0.5 for both methods.

\begin{table}[t]
\centering
\scriptsize
\setlength{\tabcolsep}{5pt}
\renewcommand{\arraystretch}{1.05}
\resizebox{0.78\columnwidth}{!}{
\begin{tabular}{lc}
\toprule
\textbf{Hyperparameter} & \textbf{Value} \\
\midrule
Training backend & FSDP2 \\
Number of GPUs & 2 \\
Precision & bf16 \\
Maximum sequence length & 4096 \\
Global batch size & 36 \\
Micro-batch size per GPU & 3 \\
Gradient accumulation steps & 6 \\
Maximum training steps & 300 \\
Learning rate & \(1\times10^{-5}\) \\
Warmup ratio & 0.1 \\
Gradient checkpointing & enabled \\
KD loss weight & 0.5 \\
\bottomrule
\end{tabular}
}
\caption{Training hyperparameters for baselines.}
\label{tab:shared-training-hparams}
\end{table}

\subsection{ACTD Training Details}
\label{app:actd-training-details}

We provide the implementation details of ACTD in
Table~\ref{tab:implementation-hparams}. To ensure a fair comparison with the baseline
methods, ACTD is trained under the same general experimental setting whenever
applicable. We implement ACTD by modifying DistillKit\footnote{\url{https://github.com/arcee-ai/DistillKit}} and fine-tune all student-model
parameters rather than using parameter-efficient adaptation.

ACTD is trained on two GPUs with bfloat16 precision. We use a per-device batch size of
3 and accumulate gradients for 6 steps, resulting in an effective batch size of 36.
The model is optimized for 300 steps with a learning rate of \(1.0\times10^{-5}\), a
cosine learning-rate schedule, and 30 warmup steps. For distillation, both the teacher
and student temperatures are set to 1.0.

\begin{table}[t]
\centering
\small
\resizebox{\columnwidth}{!}{
\begin{tabular}{lc}
\toprule
\textbf{Hyperparameter} & \textbf{Value} \\
\midrule
Fine-tuning strategy & Full-parameter \\
Number of GPUs & 2 \\
Precision & bf16 \\
Maximum sequence length & 4096 \\
Per-device batch size & 3 \\
Gradient accumulation steps & 6 \\
Effective batch size & 36 \\
Maximum training steps & 300 \\
Learning rate & \(1.0\times10^{-5}\) \\
LR scheduler & cosine \\
Optimizer & paged AdamW 8-bit \\
Weight decay & 0.025 \\
Warmup steps & 30 \\
Maximum gradient norm & 0.5 \\
Teacher temperature & 1.0 \\
Student temperature & 1.0 \\
\bottomrule
\end{tabular}
}
\caption{Training hyperparameters used for ACTD.}
\label{tab:implementation-hparams}
\end{table}

\section{Experimental Results on Other Tasks}

To evaluate the generalizability of ACTD beyond mathematical reasoning, we further conduct experiments on code generation. We use OpenCodeInstruct as the training dataset and evaluate the distilled models on HumanEval+ using Pass@1. Specifically, we distill DeepSeek-R1-Distill-Llama-8B into Qwen3-1.7B-Base, representing a cross-tokenizer distillation setting involving different model families.

As shown in Table~\ref{tab:code_generation}, ACTD achieves a Pass@1 score of 23.2, outperforming ULD, the strongest cross-tokenizer baseline, by 1.9 absolute points. ACTD also improves upon the original Qwen3-1.7B-Base student model by 2.5 points. These results demonstrate that ACTD generalizes beyond mathematical reasoning and effectively transfers code-generation capabilities across models with different tokenizers.

\begin{table}[t]
\centering
\small
\setlength{\tabcolsep}{8pt}
\begin{tabular}{lc}
\toprule
\textbf{Method} & \textbf{HumanEval+ Pass@1} \\
\midrule
Qwen3-1.7B-Base & 20.7 \\
ALM             & 15.9 \\
DSKD            & 15.9 \\
ULD             & 21.3 \\
\textbf{ACTD}   & \textbf{23.2} \\
\bottomrule
\end{tabular}
\caption{Code-generation results under the DeepSeek-R1-Distill-Llama-8B $\rightarrow$ Qwen3-1.7B-Base setting. All distilled models are trained on OpenCodeInstruct and evaluated on HumanEval+ using Pass@1.}
\label{tab:code_generation}
\end{table}

\section{The Use of Large Language Models}

We commit to accurately reporting our use of LLMs. Specifically, LLMs are used for language polishing and to improve writing clarity. However, we do not use LLMs to generate the core ideas or to write substantive portions of this paper.

\begin{figure}[t]
\centering
\small
\begin{minipage}{0.98\columnwidth}

\begin{databox}{Raw Instruction}
Return your final response within
\(\backslash\mathrm{boxed}\{\}\).
Find the center of gravity of a plate bounded by the lines
\(x=0\), \(y=0\),
\(\frac{x^{2}}{a^{2}}+\frac{y^{2}}{b^{2}}=1\),
\(y \geqslant 0\), and \(x \geqslant 0\), given that the surface
density is \(P=xy\).
\end{databox}

\vspace{4pt}

\begin{databox}{Original Completion}
To find the coordinates of the center of mass, we use
\[
x_C=\frac{m_y}{M},\qquad
y_C=\frac{m_x}{M}.
\]
The mass and moments are computed over the first-quadrant ellipse.

\(\cdots\)

Evaluating the moment about the \(x\)-axis gives
\[
m_x=\frac{a^2b^3}{15}.
\]

4. Determine the coordinates:
\[
x_C
=\frac{m_y}{M}
=\frac{\frac{a^3b^2}{15}}
       {\frac{a^2b^2}{8}}
=\frac{8a}{15},
\]
\[
y_C
=\frac{m_x}{M}
=\frac{\frac{a^2b^3}{15}}
       {\frac{a^2b^2}{8}}
=\frac{8b}{15}.
\]

Thus, the center of gravity is
\[
\boxed{\left(\frac{8a}{15},\frac{8b}{15}\right)}.
\]
\end{databox}

\vspace{4pt}

\begin{databox}{Prefix-Truncated Completion}
To find the coordinates of the center of mass, we use
\[
x_C=\frac{m_y}{M},\qquad
y_C=\frac{m_x}{M}.
\]
The mass and moments are computed over the first-quadrant ellipse.

\(\cdots\)

Evaluating the moment about the \(x\)-axis gives
\[
m_x=\frac{a^2b^3}{15}.
\]

4. Determine the coordinates \((x_C,y_C)\):

\noindent
\textcolor{blue}{\texttt{\textless/think\textgreater}}\par

\[
\boxed{\left(\frac{8a}{15},\frac{8b}{15}\right)}
\]
\end{databox}

\end{minipage}

\caption{
An example of prefix truncation. The original reasoning is truncated
before the coordinate calculation, after which a forced
\textcolor{blue}{\texttt{\textless/think\textgreater}} marker and the
preserved final-answer suffix are appended.
}
\label{fig:prefix_truncation_example}
\end{figure}

\end{document}